\documentclass[11pt,a4paper]{article}
\usepackage[hyperref]{emnlp-ijcnlp-2019}
\usepackage{times}
\usepackage{latexsym}

\usepackage{url}
\usepackage{times}
\usepackage{latexsym}
\usepackage{graphicx}
\usepackage{caption}
\usepackage{subcaption}
\usepackage{amsmath}
\usepackage{multirow}
\usepackage{booktabs}
\usepackage{forest}
\usepackage{float}
\aclfinalcopy 

\usepackage[]{todonotes}  

\DeclareMathOperator*{\argmax}{argmax}

\title{Do Attention Heads in BERT Track Syntactic Dependencies?}

\author{Phu Mon Htut\thanks{\ \ Equal contribution}~\,$^1$ \\\And
Jason Phang\footnotemark[1]~\hspace{0.2em}$^{1}$  \\\And
 Shikha Bordia\footnotemark[1]~\hspace{0.2em}\thanks{\ \ Currently working at Verisk Analytics. This work was completed when the author was at New York University.}~\hspace{0.2em}$^{2}$
 \\\And
  Samuel R. Bowman$^{1,2,3}$  
 \AND
$^{1}$\normalfont Center for Data Science\\New York University\\60 Fifth Avenue\\New York, NY 10011\And
$^{2}$\normalfont Dept. of Computer Science\\New York University\\60 Fifth Avenue\\New York, NY 10011\And
$^{3}$\normalfont Dept. of Linguistics\\New York University\\10 Washington Place\\New York, NY 10003
  }

\date{}

\begin{document}
\maketitle
\begin{abstract}

We investigate the extent to which individual attention heads in pretrained transformer language models, such as BERT and RoBERTa, implicitly capture syntactic dependency relations.
We employ two methods---taking the maximum attention weight and computing the maximum spanning tree---to extract implicit dependency relations from the attention weights of each layer/head, and compare them to the ground-truth Universal Dependency (UD) trees. 
We show that, for some UD relation types, there exist heads that can recover the dependency type significantly better than baselines on parsed English text, suggesting that some self-attention heads act as a proxy for syntactic structure. 
We also analyze BERT fine-tuned on two datasets---the syntax-oriented CoLA and the semantics-oriented MNLI---to investigate whether fine-tuning affects the patterns of their self-attention, but we do not observe substantial differences in the overall dependency relations extracted using our methods. 
Our results suggest that these models have some specialist attention heads  that track individual dependency types, but no generalist head that performs holistic parsing significantly better than a trivial baseline, and that analyzing attention weights directly may not reveal much of the syntactic knowledge that BERT-style models are known to learn.
\end{abstract}

\section{Introduction}
    
%
    

Pretrained Transformer models such as BERT \citep{devlin2018bert} and RoBERTa \citep{liu2019roberta} have shown stellar performance on language understanding tasks, significantly improve the state-of-the-art on many tasks such as dependency parsing \citep{Zhou19LIMITBERT}, question answering \citep{Rajpurkar2016}, and have attained top positions on transfer learning benchmarks such as GLUE \cite{wang-etal-2018-glue} and SuperGLUE \citep{WangSuperGlue}. As these models become a staple component of many NLP tasks, it is crucial to understand what kind of knowledge they learn, and why and when they perform well. To that end, researchers have investigated the linguistic knowledge that these models learn by analyzing BERT \citep{GoldbergBert2019,lin-etal-2019-open} directly or training probing classifiers on the contextualized embeddings or attention heads of BERT \citep{tenney2018what,tenney-etal-2019-bert,hewitt2019structural}. 

BERT and RoBERTa, as Transformer models \citep{vaswani2017attention}, compute the hidden representation of all the attention heads at each layer for each token by attending to all the token representations in the preceding layer. In this work, we investigate the hypothesis that BERT-style models use at least some of their attention heads to track syntactic dependency relationships between words. We use two dependency relation extraction methods to extract dependency relations from each self-attention heads of BERT and RoBERTa. The first method---maximum attention weight (\textsc{Max})---designates the word with the highest incoming attention weight as the parent, and is meant to identify specialist heads that track specific dependencies like \texttt{obj} \citep[in the style of][]{clark-etal-2019-bert}. The second---maximum spanning tree (MST)---computes a maximum spanning tree over the attention matrix, and is meant to identify generalist heads that can form complete, syntactically informative dependency trees. We analyze the extracted dependency relations and trees to investigate whether the attention heads of these models track syntactic dependencies significantly better than chance or baselines, and what type of dependency relations they learn best. 
In contrast to probing models \citep{Adi2017,conneau-etal-2018-cram}, our methods require no further training.  
In prior work, \citet{clark-etal-2019-bert} find that some heads of BERT exhibit the behavior of some dependency relation types, though they do not perform well at all types of relations in general. We are able to replicate their results on BERT using our \textsc{Max} method. In addition, we also perform a similar analysis on BERT models fine-tuned on natural language understanding tasks as well as RoBERTa. 

Our experiments suggest that there are particular attention heads of BERT and RoBERTa that encode certain dependency relation types such as \texttt{nsubj}, \texttt{obj} with substantially higher accuracy than our baselines---a randomly initialized Transformer and relative positional baselines. 
We find that fine-tuning BERT on the syntax-oriented CoLA does not significantly impact the accuracy of extracted dependency relations. However, when fine-tuned on the semantics-oriented MNLI dataset, we see improvements in accuracy for longer-distance clausal relations and a slight loss in accuracy for shorter-distance relations. Overall, while BERT and RoBERTa models obtain non-trivial accuracy for some dependency types such as \texttt{nsubj}, \texttt{obj} and \texttt{conj} when we analyze individual heads, their performance still leaves much to be desired. 
On the other hand, when we use the MST method to extract full trees from specific dependency heads, BERT and RoBERTa fail to meaningfully outperform our baselines. Although the attention heads of BERT and RoBERTa capture several specific dependency relation types somewhat well, they do not reflect the full extent of the significant amount of syntactic knowledge that these models are known to learn.

\section{Related Work}
Previous works have proposed methods for extracting dependency relations and trees from the attention heads of the transformer-based neural machine translation (NMT) models. In their preliminary work, \citet{marecek-rosa-2018-extracting} aggregate the attention weights across the self-attention layers and heads to form a single attention weight matrix. Using this matrix, they propose a method to extract constituency and (undirected) dependency trees by recursively splitting and constructing the maximum spanning trees respectively. In contrast, \citet{raganato-tiedemann-2018-analysis} train a Transformer-based machine translation model on different language pairs and extract the maximum spanning tree algorithm from the attention weights of the encoder for each layer and head individually. They find that the best dependency score is not significantly higher than a right-branching tree baseline. \citet{voita-etal:2019a:ACL} find the most confident attention heads of the Transformer NMT encoder based on a heuristic of the concentration of attention weights on a single token, and find that these heads mostly attend to relative positions, syntactic relations, and rare words. 

Additionally, researchers have investigated the syntactic knowledge that BERT learns by analyzing the contextualized embeddings \citep{warstadt-etal-2019-investigating} and attention heads of BERT \citep{clark-etal-2019-bert}. \citet{GoldbergBert2019} analyzes the contextualized embeddings of BERT by computing language model surprisal on subject-verb agreement and shows that BERT learns significant knowledge of syntax. \citet{tenney2018what} introduce a probing classifier for evaluating syntactic knowledge in BERT and show that BERT encodes syntax more than semantics. \citet{hewitt2019structural} train a \textit{structural probing} model that maps the hidden representations of each token to an inner-product space that corresponds to syntax tree distance. They show that the learned spaces of strong models such as BERT and ELMo \citep{peters2018elmo} are better for reconstructing dependency trees compared to baselines. \citet{clark-etal-2019-bert} train a probing classifier on the attention-heads of BERT and show that BERT's attention heads capture substantial syntactic information. While there has been prior work on analysis of the attention heads of BERT, we believe we are the first to analyze the dependency relations learned by the attention heads of fine-tuned BERT models and RoBERTa. 



\section{Methods}
\label{sec:methods}

 


\subsection{Models}

BERT \citep{devlin2018bert} is a Transformer-based masked language model, pretrained on BooksCorpus \citep{bookcorpus} and English Wikipedia, that has attained stellar performance on a variety of downstream NLP tasks. RoBERTa \citep{liu2019roberta} adds several refinements to BERT while using the same model architecture and capacity, including a longer training schedule over more data, and shows significant improvements over BERT on a wide range of NLP tasks. We run our experiments on the pretrained large versions of both BERT (cased and uncased) and RoBERTa models, which consist of 24 self-attention layers with 16 heads each layer. For a given dataset, we feed each input sentence through the respective model and capture the attention weights for each individual head and layer. 

\citet{phang2018stilts} report the performance gains on the GLUE benchmark by supplementing pretrained BERT with data-rich supervised tasks such as the Multi-Genre Natural Language Inference dataset \citep[MNLI; ][]{WilliamsNB17}. Although these fine-tuned BERT models may learn different aspects of language and show different performance from BERT on GLUE benchmark, comparatively little previous work has investigated the syntax learned by these fine-tuned models \citep{warstadt-etal-2019-investigating}. We run experiments on the uncased BERT-large model fine-tuned on the Corpus of Linguistic Acceptability \citep[CoLA;][]{warstadt2018neural} and MNLI to investigate the impact of fine-tuning on a syntax-related task (CoLA) and a semantic-related task (MNLI) on the structure of attention weights and resultant extracted dependency relations. We refer to these fine-tuned models as \textit{CoLA-BERT} and \textit{MNLI-BERT} respectively.

\subsection{Analysis Methods}

We aim to test the hypothesis that the attention heads of BERT learn syntactic relations implicitly, and that the self-attention between two words corresponds to their dependency relation. We use two methods for extracting dependency relations from the attention heads of Transformer-based models. Both methods operate on the attention weight matrix $W\in(0,1)^{T\times T}$ for a given head at a given layer, where $T$ is the number of tokens in the sequence, and the rows and columns correspond to the attending and attended tokens respectively (such that each row sums to 1). 

\paragraph{Method 1: Maximum Attention Weights (\textsc{Max})}  Given a token A in a sentence, a self-attention mechanism is designed to assign high attention weights on tokens that have some kind of relationship with token A \citep{vaswani2017attention}. Therefore, for a given token A, a token B that has the highest attention weight with respect to the token A should be related to token A. Our aim is to investigate whether this relation maps to a universal dependency relation. We assign a relation ($w_i$, $w_j$) between word $w_i$ and $w_j$ if $j=\argmax \;W[i]$ for each row (that corresponds to a word in attention matrix) $i$ in attention matrix $W$. Based on this simple strategy, we extract relations for all sentences in our evaluation datasets. This method is similar to \citet{clark-etal-2019-bert}, and attempts to recover individual arcs between words; the relations extracted using this method need not form a valid tree, or even be fully connected, and the resulting edge directions may or may not match the canonical directions. Hence, we evaluate the resulting arcs individually and ignore their direction. After extracting dependency relations from all heads at all layers, we take the maximum UUAS over all relations types.

\paragraph{Method 2: Maximum Spanning Tree (MST)} We also want to investigate if there are attention heads of BERT that can form complete, syntactically informative parse trees. To extract complete valid dependency trees from the attention weights for a given layer and head, we follow the approach of \citet{raganato-tiedemann-2018-analysis} and treat the matrix of attention weight tokens as a complete weighted directed graph, with the edges pointing from the output token to each attended token. As in \citeauthor{raganato-tiedemann-2018-analysis}, we take the root of the gold dependency tree as the starting node and apply the Chu-Liu-Edmonds algorithm \citep{liu1965shortest,edmonds1967optimum} to compute the maximum spanning tree. (Using the gold root as the starting point in MST may artificially improve our results slightly, but this bias is applied evenly across all the models we compare.) The resulting tree is a valid directed dependency tree, though we follow \citet{hewitt2019structural} in evaluating it as undirected, for easier comparison with our \textsc{Max} method. 

\paragraph{}Following \citet{voita-etal:2019a:ACL}, we exclude the sentence demarcation tokens (\texttt{[CLS]}, \texttt{[SEP]}, \texttt{<s>}, \texttt{</s>}) from the attention matrices. This allows us to focus on inter-word attention. Where the tokenization of our parsed corpus does not match the model's tokenization, we merge the non-matching tokens until the tokenizations are mutually compatible, and sum the attention weights for the corresponding columns and rows. We then apply either of the two extraction methods to the attention matrix. During evaluation when we compare the gold dependencies, to handle the subtokens within the merged tokens, we set all subtokens except for the first to depend on the first subtoken. This approach is largely similar to that in \citet{hewitt2019structural}. We use the English Parallel Universal Dependencies (PUD) treebank from the CoNLL 2017 shared task \citep{PUDZemanHPPSGNP18} as the gold standard for our evaluation.

\begin{figure}
    \includegraphics[width=\columnwidth] {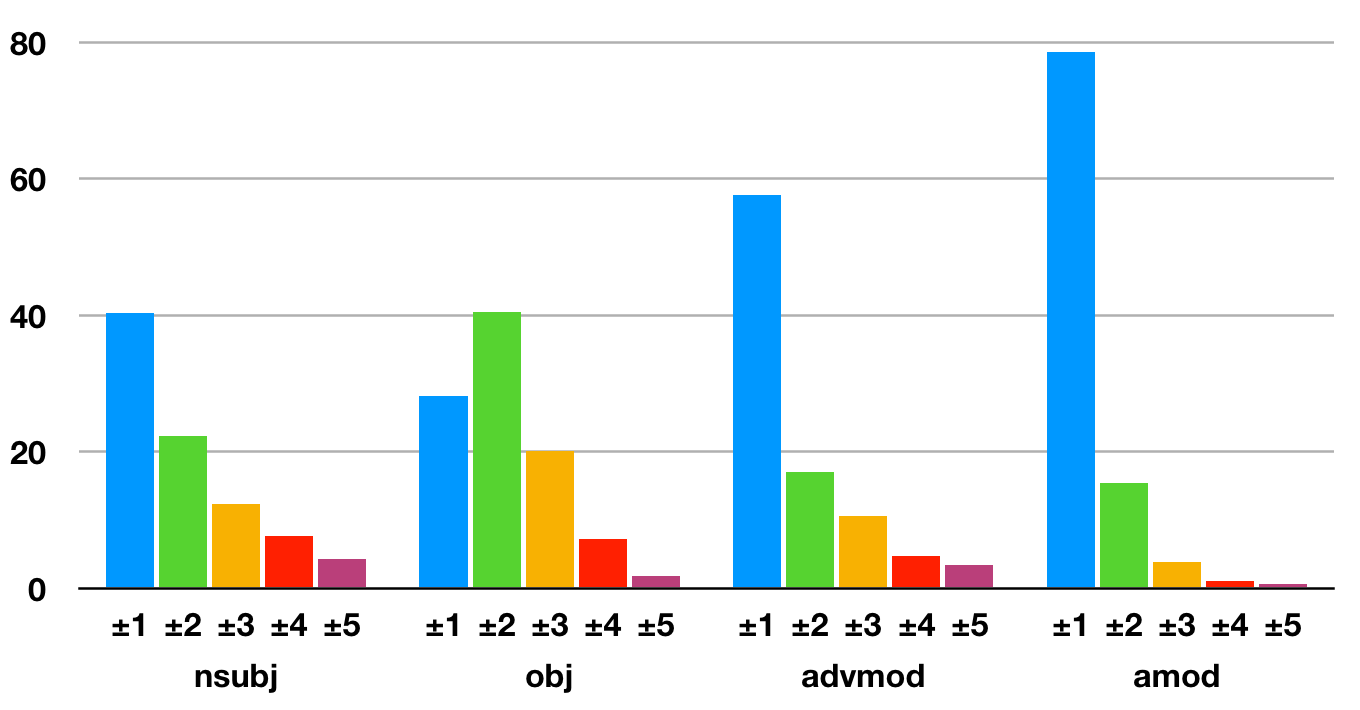}
    \caption{The distribution of the relative positions of \texttt{nsubj}, \texttt{obj}, \texttt{advmod}, \texttt{amod} in PUD.}
    \label{fig:relative_pos_pud}
\end{figure}

\begin{table*}[t!]
\small
\centering
\setlength{\tabcolsep}{3 pt} 
\begin{center}
\begin{tabular}{lrrrrrrrrrrrrrrrr}
\toprule
   \bf Model  & \bf nsubj &  \bf obj & \bf advmod & \bf amod  & \bf case &  \bf det & \bf obl & \bf nmod  & \bf punct & \bf aux & \bf conj  & \bf cc & \bf mark   & & \bf advcl & \bf csubj \\
 \midrule
BERT (cased)  & \textit{56.8} & \textit{77.4} & \textit{59.6} & \textit{82.7} & \textit{ 83.8} &  \textit{93.2} &  \textit{31.0} &  \textit{51.7} &  \textit{40.2} & \textit{80.0} &  \textit{45.0}  & \bf \textit{73.5} & \bf \textit{ 72.1}  &  & \textit{29.4} &  \textit{55.6} \\
BERT   & \bf \textit{60.7} & \bf \textit{86.4} & \bf \textit{62.6} &  \textit{84.0} & \textit{88.8} &   \bf \textit{96.3} &  \textit{32.1} & \bf \textit{66.7}  & \textit{40.5} & \textit{81.0}   &  \textit{48.9}  &  \textit{67.4} &  \bf \textit{72.1}  &  & \textit{26.6} &  \textit{48.1}  \\
 CoLA-BERT  & \textit{59.7} & \textit{84.2} & \textit{63.8} & \bf \textit{86.4} &   \bf  \textit{89.5} &   \textit{96.2} &  \bf \textit{33.8} &  \textit{66.2}  & \textit{41.1} &  \bf  \textit{82.2} & \bf  \textit{50.6} &  \textit{68.5} &  \textit{70.8}  &   & \textit{28.0} &  \textit{51.9} \\
 MNLI-BERT  & \textit{59.5} & \textit{83.0} & \textit{59.7} & \textit{81.7} &   \textit{87.9} &   \textit{95.3} &  \textit{32.4} &  \textit{63.3}  & \bf  \textit{41.3} & \textit{78.6} &  \textit{50.5}  &  \textit{65.5} &  \textit{68.5}  &   & \bf \textit{34.5} &  \bf \textit{63.0} \\
 RoBERTa  & \textit{50.2} & \textit{69.3} & \textit{58.5} & \textit{79.3} &   \textit{75.6} &  \textit{74.4} & \textit{26.2} & \textit{47.4}  & \textit{37.4} & \textit{75.7} &  \textit{44.6}  & \textit{69.0} & \textit{63.1}  &   & \textit{23.2} & \textit{44.4} \\
Positional  & 40.4 & 40.5  & 57.6 & 78.7  & 38.7 &  56.7 & 24.0 & 35.4  & 18.6 & 55.5 & 27.8  & 43.4 & 53.7  &  &  10.23 & 25.9 \\
  Random-BERT & 16.8 & 12.9 & 11.8 &  11.1 & 13.7 &  12.6 & 13.5 & 13.8  & 12.6 & 16.3 & 18.9  & 20.9 & 12.8  &  & 13.3 & 22.2 \\
\bottomrule 
\end{tabular}
\end{center}
\caption{\label{tab:top-type-accuracies} Highest accuracy for the most frequent dependency types. 
\textbf{Bold} marks the highest accuracy for each dependency type based on our \textsc{Max} method. \textit{Italics} marks accuracies that outperform our trivial baselines. 
}
\end{table*}

\paragraph{Baselines} Many dependency relations tend to occur in specific positions relative to the parent word. For example, \texttt{amod} (adjectival modifier) mostly occurs before a noun. As an example, Figure~\ref{fig:relative_pos_pud} shows the distribution of relative positions for four major UD relations in our data. Following \citet{voita-etal:2019a:ACL}, we compute the most common positional offset between a parent and child word for a given dependency relation, and formulate a baseline based on that most common relative positional offset to evaluate our methods. For MST, as we also want to evaluate the quality of the entire tree, we use a right-branching dependency tree as baseline.
Additionally, we use a BERT-large model with randomly initialized weights (which we refer to as \textit{random BERT}), as previous work has shown that randomly initialized sentence encoders perform surprisingly well on a suite of NLP tasks \citep{Zhang2018LM,Wieting2019NoTraining}.

\section{Results}
\begin{figure}[t]
    \centering
        \includegraphics[width=\columnwidth] {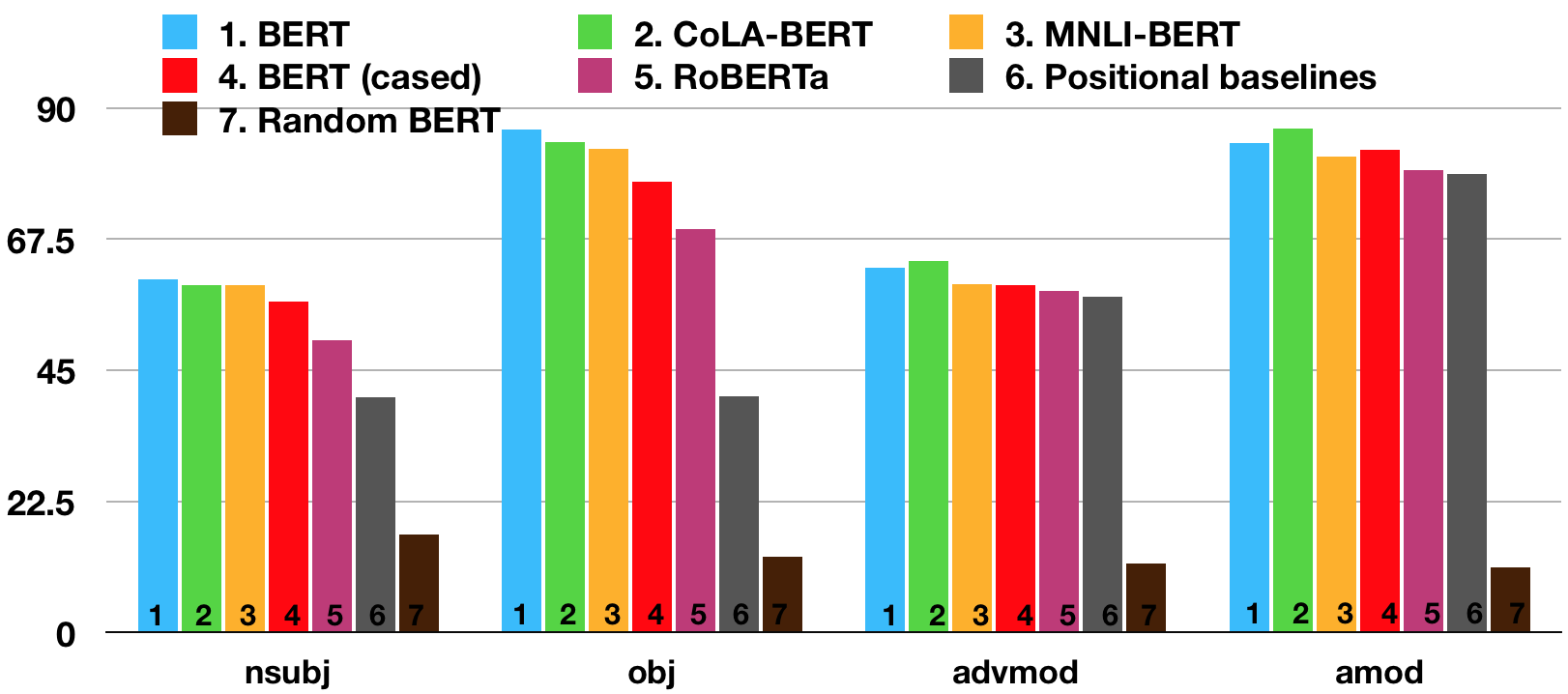}
    \caption{ Undirected dependency accuracies by type based on our \textsc{Max} method.}
    \label{fig:pud-accuracies}
\end{figure}

Figure~\ref{fig:pud-accuracies} and Table \ref{tab:top-type-accuracies} describe the accuracy for the  most frequent relation types in the dataset using relations extracted based on the \textsc{Max} method. We also include results for the rarer long-distance \texttt{advcl} and \texttt{csubj} dependency types, as they show that MNLI-BERT has a tendency to track clausal dependencies better than BERT, CoLA-BERT, and RoBERTa. The non-random models outperform random BERT substantially for all dependency types. They also outperform the relative position baselines for some relation types. They outperform the baselines by a large margin for \texttt{nsubj} and \texttt{obj}, but only slightly better for \texttt{advmod} and \texttt{amod}. These results are consistent with the findings of \citet{clark-etal-2019-bert}. Moreover, we do not observe substantial differences in accuracy by fine-tuning on CoLA. Both BERT and CoLA-BERT have similar or slightly better performance than MNLI-BERT, except for clausal dependencies such as \texttt{advcl} (adverbial clause modifier) and \texttt{csubj} (clausal subject) where MNLI-BERT outperforms BERT and CoLA-BERT by more than 5 absolute points in accuracy. This suggests that fine-tuning on a semantics-oriented task encourages effective long-distance dependencies, although it slightly degrades the performance in other shorter-distance dependency types.

\begin{figure}[t]
    \centering
        \includegraphics[width=\columnwidth] {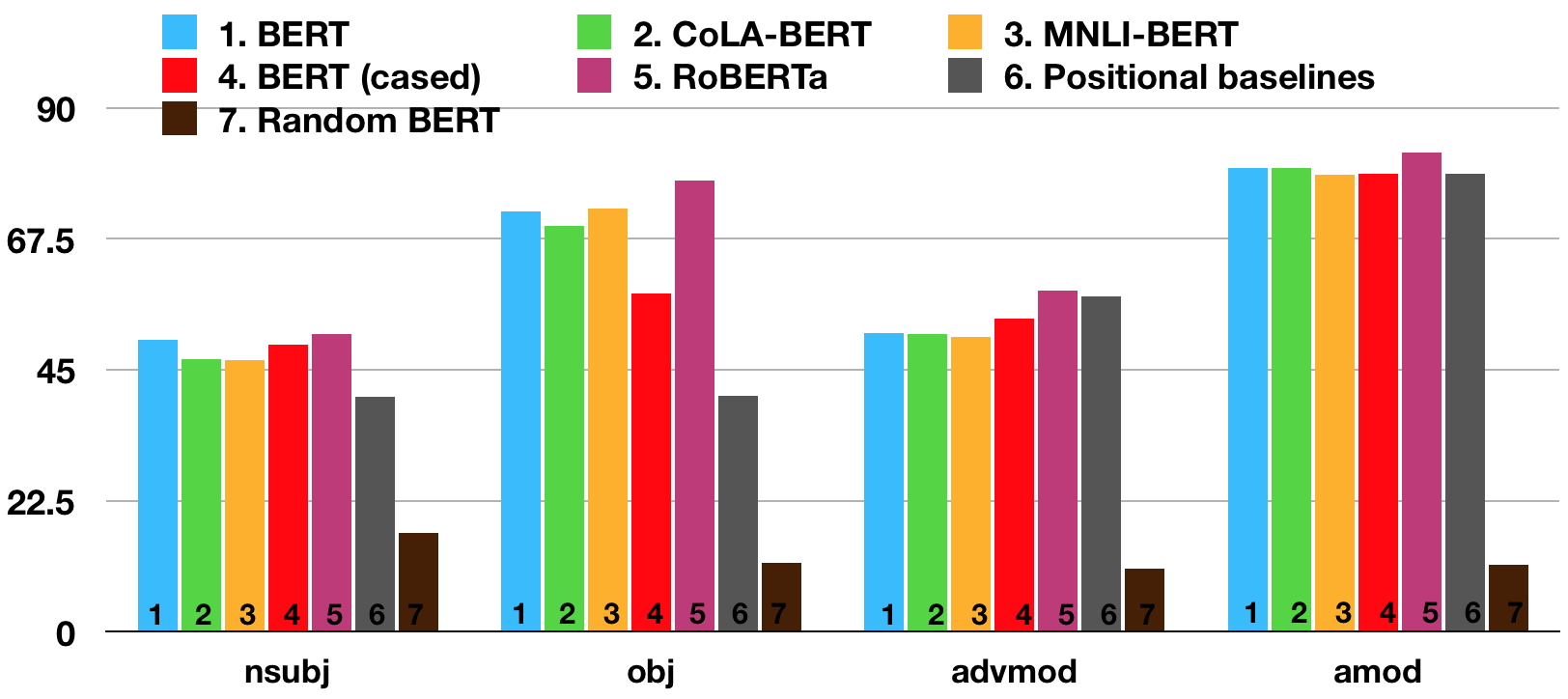}
    \caption{ Undirected dependency accuracies by type based on the MST method.}
    \label{fig:mst-accuracies}
\end{figure}

Figure \ref{fig:mst-accuracies} shows the accuracy for \texttt{nsubj, obj, advmod}, and \texttt{amod} relations extracted based on the MST method. Similar to the \textsc{Max} method, we choose the best accuracy for each relation type. We observe that the models outperform the baselines by a large margin for \texttt{nsubj} and \texttt{obj}. However, the models do not outperform the positional baseline for \texttt{advmod} and \texttt{amod}. Surprisingly, RoBERTa performs worse than other BERT models in all categories when the \textsc{Max}  method is used to extract the trees, but it outperforms all other models when the MST method is used.

\begin{figure}[t]
    \centering
        \includegraphics[width=\columnwidth] {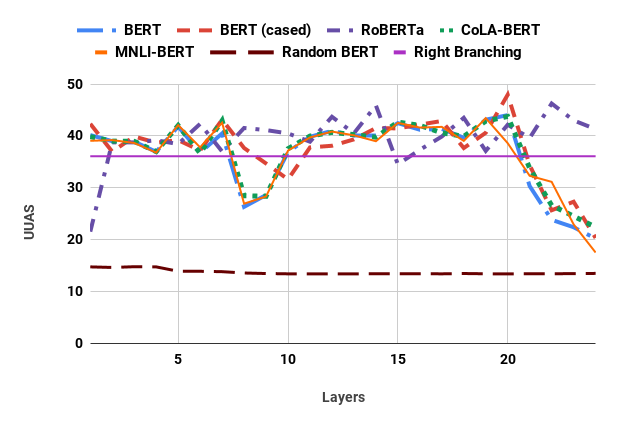}
    \caption{Maximum UUAS across layers of dependency trees extracted based on the MST method on PUD. 
    }
    \label{fig:uuas}
\end{figure}

Figure~\ref{fig:uuas} describes the maximum undirected unlabeled attachment scores (UUAS) across each layer. The trained  models achieve significantly higher UUAS than random BERT.  Although the trained models perform better than the right-branching baseline in most cases, the performance gap is not substantial. Given that the MST method uses the root of the gold trees, whereas the right-branching baseline does not, 
this implies that the attention weights in the different layers/heads of the BERT models do not appear to correspond to complete, syntactically informative parse trees.

Overall, the results of both analysis methods suggest that, although some attention heads of BERT capture specific dependency relation types, they do not reflect the full extent of the significant amount of syntactic knowledge BERT and RoBERTa are known to learn as shown in previous syntactic probing work \citep{tenney2018what,hewitt2019structural}. Additionally, we find that fine-tuning on the semantics-oriented MNLI dataset improves long term dependencies while slightly degrading the performance for other dependency types. The overall performance of BERT and the fine-tuned BERTs over the non-random baselines are not substantial, and fine-tuning on CoLA and MNLI also does not have a large impact on UUAS.

\section{Conclusion}
In this work, we investigate whether the attention heads of BERT and RoBERTa exhibit the implicit syntax dependency by extracting and analyzing the dependency relations from the attention heads at all layers. We use two simple dependency relation extraction methods that require no additional training, and observe that there are certain specialist attention heads of the models that track specific dependency types, but neither of our analysis methods support the existence of generalist heads that can perform holistic parsing. Furthermore, we observe that fine-tuning on CoLA and MNLI does not significantly change the overall pattern of self-attention within the frame of our analysis, despite their being tuned for starkly different downstream tasks.

\section*{Acknowledgments}

This project grew out of a class project for the Spring 2019 NYU Linguistics seminar \textit{Linguistic Knowledge in Reusable Sentence Encoders}. We are grateful to the department for making this seminar possible.

This material is based upon work supported by the National Science Foundation under Grant No. 1850208. Any opinions, findings, and conclusions or recommendations expressed in this material are those of the author(s) and do not necessarily reflect the views of the National Science Foundation. This project has also benefited from financial support to SB by Samsung Research under the project \textit{Improving Deep Learning using Latent Structure}
and from the donation of a Titan V GPU by NVIDIA Corporation. 
\bibliography{emnlp-ijcnlp-2019}
\bibliographystyle{acl_natbib}


\end{document}